\documentclass[letterpaper, 10 pt, conference]{ieeeconf}
\IEEEoverridecommandlockouts
\usepackage[utf8]{inputenc}

\usepackage{balance} 
\usepackage{color}
\usepackage{soul,xcolor}
\usepackage{amsmath} 
\usepackage{graphicx}
\usepackage{comment}
\usepackage[nodisplayskipstretch]{setspace}
\usepackage{booktabs}
\usepackage[ruled, vlined, noend, linesnumbered]{algorithm2e}
\usepackage[noadjust]{cite} 
\usepackage{amssymb}
\usepackage{array}\usepackage{makecell}

\usepackage{indentfirst}
\usepackage{color}
\usepackage{soul,xcolor}
\usepackage{amsmath} 
\usepackage{graphicx}
\usepackage{subcaption}
\usepackage{caption,subcaption}
\usepackage{multicol}
\usepackage[normalem]{ulem}
\usepackage{balance}
\usepackage[ruled, vlined, noend, linesnumbered]{algorithm2e}
\usepackage{lipsum}
\usepackage{algorithmic}

\usepackage{wrapfig}

\usepackage{bibentry}
\usepackage{hyperref}

 \newcommand{\algo}{IODA}
 \newcommand{\algofull}{Imaginary Out-of-Distribution Actions}

\title{\LARGE \bf
 Imagining In-distribution States: How Predictable Robot Behavior Can Enable User Control Over Learned Policies
}

\author{Isaac Sheidlower$^{1}$, Emma Bethel$^{1}$, Douglas Lilly$^{1}$, Reuben M. Aronson$^{1}$, and Elaine Schaertl Short$^{1}$ 
\thanks{$^{1}$ Tufts University, Medford, MA 02144, USA}%
}

\begin{document}

\maketitle
\thispagestyle{empty}
\pagestyle{empty}

\begin{abstract}
It is crucial that users are empowered to take advantage of the functionality of a robot and use their understanding of that functionality to perform novel and creative tasks. Given a robot trained with Reinforcement Learning (RL), a user may wish to leverage that autonomy along with their familiarity of how they expect the robot to behave to collaborate with the robot. One technique is for the user to take control of some of the robot's action space through teleoperation, allowing the RL policy to simultaneously control the rest. We formalize this type of shared control as Partitioned Control (PC). However, this may not be possible using an out-of-the-box RL policy. For example, a user's control may bring the robot into a failure state from the policy's perspective, causing it to act unexpectedly and hindering the success of the user's desired task. In this work, we formalize this problem and present \algofull{}, \algo{}, an initial algorithm which empowers users to leverage their expectations of a robot's behavior to accomplish new tasks. We deploy IODA in a user study with a real robot and find that IODA leads to both better task performance and a higher degree of alignment between robot behavior and user expectation. We also show that in PC, there is a strong and significant correlation between task performance and the robot's ability to meet user expectations, highlighting the need for approaches like IODA. Code is available at \url{https://github.com/AABL-Lab/ioda_roman_2024}
\end{abstract}

\section{Introduction}
\vspace{-1mm}
As robots become more prevalent among experts and non-experts alike, it is necessary to ensure that people are empowered to use robots and their functionality to accomplish their desired tasks. Ideally, a robot can be deployed with an exhaustive library of Reinforcement Learning (RL) policies that can autonomously perform any task the user requests. However, this approach is impractical: users may wish to use the robot for a task or in an environment that was not foreseen by the designers of the robot or for a task that a user may have difficulty specifying. Realistically, the user will be provided with a limited set of policies and ways of controlling the robot, such as teleoperation. Therefore, it is essential to develop methods that allow users to take advantage both types of control to accomplish novel and unforeseen tasks.  

Consider the following example: Sally often uses her assistive robot arm to perform tasks in and around the house. She has used the robot's pick-and-place RL policy to move cups of liquid, such as coffee or tea, many times. Because of this, she expects that the robot will pick up the cup and steadily move it to a specified location. Now she wants to use the same functionality to water her flowers. To do this, she requests for the robot to autonomously carry a cup of water over her flower bed and during execution rotates the robot's wrist to pour the water over her flowers as the robot moves along its path. ''As-is'', an RL policy may not facilitate Sally's goal: if it is brought out-of-distribution because of the rotated wrist, it may move erratically, impeding task success. Similar problems may also arise in autonomous driving, such as driving on the sidewalk or hitting an obstacle to avoid a more serious accident. 

\begin{figure}[t]
\centering
\includegraphics[width=.475\textwidth]{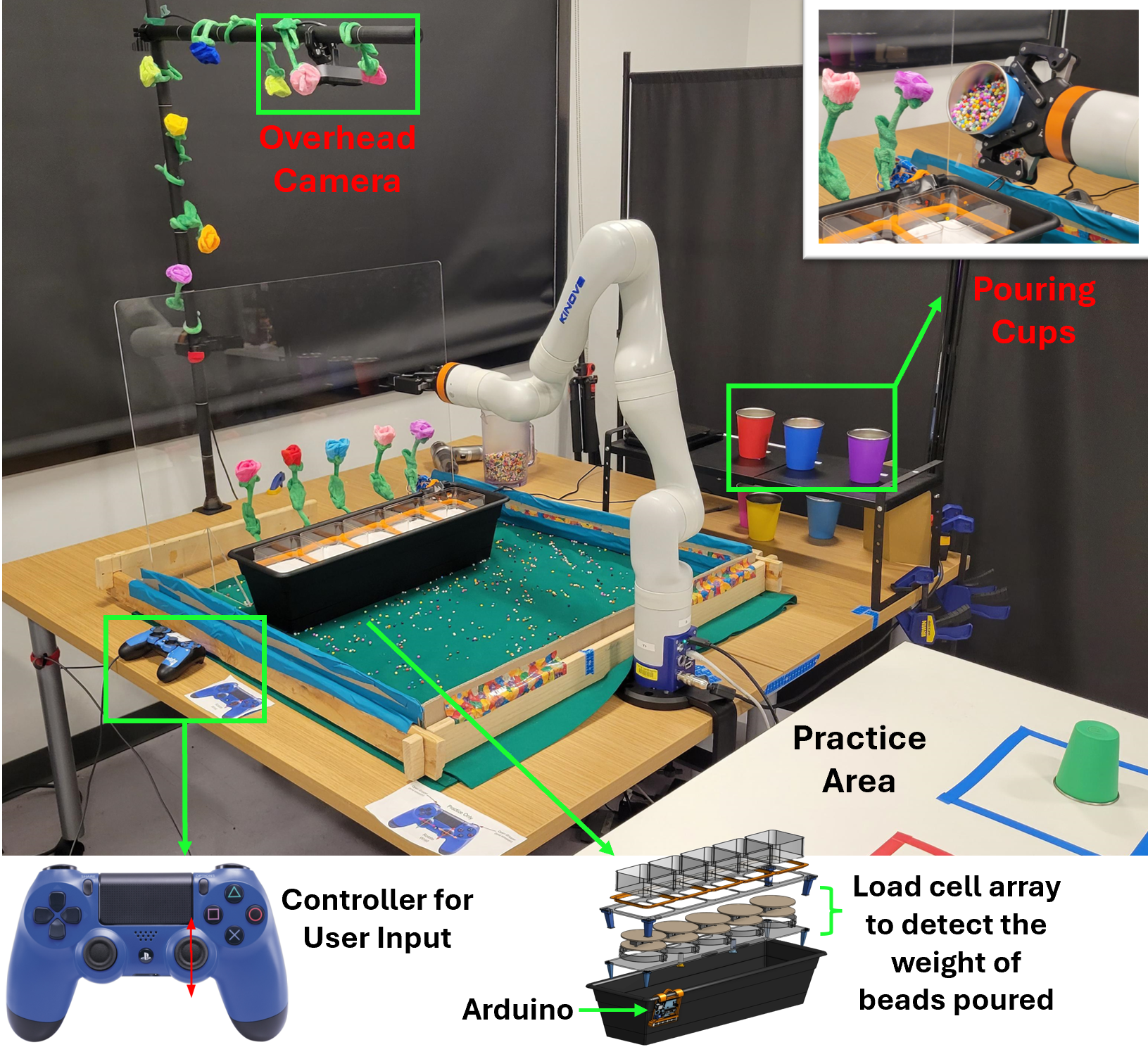}%
\captionsetup{indention=1.5em}
\captionof{figure}{A depiction of the ''flower watering" task setup used to study Partitioned Control and IODA with novice-users.}%
\label{fig:introfig}
\end{figure}

These examples demonstrate potential problems of naively using pre-trained policy specified for a given task while the user controls part of the robot's behavior. Since the user may have only seen successful examples of task execution, they may not know that their control signal could cause failure with respect to the robot's reward function or safety constraints, causing it to behave unexpectedly. Furthermore, unexpected robot behavior may impede the user's ability to perform novel tasks or result in task failure from the user's perspective. This setting also enables the user to bring the policy ``out-of-distribution,`` which may result in unwanted behavior. It is essential to study this type of interaction from the user's perspective and to create algorithms that modify RL policies to better assist the user in completing their task. We call this interaction \emph{Partitioned Control (PC)}: when the user has teleoperation over certain parts of the robot's action space while the others are fully autonomous. Our key insight is that during PC, the robot should act in a way similar to the behavior the user has seen before and is familiar with, allowing the user to leverage their prior knowledge about the robot's behavior to adapt it to new tasks.

In this paper, we formalize Partitioned Control and the aforementioned problem setting, and present the algorithm \algofull{} (\algo{}), which modifies the state passed into an RL policy by projecting its current state to a previous state the user has seen: despite the robot's current real state, it chooses actions based on an \emph{imagined} state. \algo{} only modifies the state when the user brings the robot out-of-distribution (OOD) concerning states the user is familiar with/has seen before. IODA uses an OOD detector to determine when to modify the state, and a projection function to project the state onto the nearest state familiar to the user. We demonstrated \algo{} in simulation as a proof-of-concept \cite{sheidlower_modifying_2023}. We reiterate that analysis here before deploying \algo{} in a user study (\textbf{n=18)} where we compare users using IODA to an unaltered RL policy as well as a heuristic-based approach for handling OOD states where the robot will stop moving when it is OOD (we call this condition ''STOP"). Users perform a "flower watering" task (Fig. 1.) similar to the one described previously. We find that IODA leads to both better task performance as well as a higher degree of alignment between robot behavior and user expectations. Furthermore, we show in Partitioned Control, there is a strong and significant correlation between meeting user expectations and achieving high task performance.

\section{Related Work}
\vspace{-1mm}
Reinforcement learning (RL) is a common and effective way for robots to learn new tasks \cite{kaelbling_reinforcement_1996, kober_reinforcement_nodate, ibarz_how_2021}. Much of the success of RL for robotics has come from Deep RL \cite{arulkumaran_brief_2017, nguyen_review_2019} and human-in-the-loop learning \cite{christiano_deep_2017, arzate_cruz_survey_2020, schrum_mind_2022}. However, such methods assume that the robot already knows how to perform the task the user requests or that a user is willing to spend time teaching a robot. These assumptions may limit the scope or applicability of any RL-based robot to one or a few tasks. While methods such as multitask RL \cite{xie_lifelong_2022, shridhar_perceiver-actor_2023, walke_dont_2023}, large-language model skill grounding \cite{ichter_as_2023, driess_palm-e_2023, liang_code_2023}, and behavioral-diversity learning \cite{eysenbach_diversity_2018, osa_discovering_2022} have all partially addressed this problem by enlarging and diversifying the robot's task repertoire, they do not explicitly empower users with a method of controlling or collaborating with the robot other than through task specification.

In contrast to a fully autonomous policy, Shared Control (SC) alleviates some of the burdens of teleoperation while allowing assistance from autonomous robot behavior \cite{li_continuous_2015, matsumoto_shared_2022, bustamante_toward_2021, dragan_policy-blending_2013}. SC is often approached as a blending of the user's and robot's control. This is similar to Shared Autonomy (SA) \cite{javdani_shared_2015, gopinath_human---loop_2017, selvaggio_autonomy_2021, fontaine_differentiable_2021, reddy_shared_2018}, where a user's control signal is interpreted as an indication of their desired goal. Typically in SA it is assumed that a user's goal is known in advance and can be represented in contrast to other goals in the environment. Although this assumption has been relaxed in some previous work by updating the potential set of goals \cite{zurek_situational_2021, fontaine_differentiable_2021, yoneda_noise_nodate}, these approaches require the user to first demonstrate the task or a similar task without assistance. In contrast, we focus on a user leveraging their creative problem-solving skills to partially control an otherwise fully autonomous robot to complete a novel task. The robot is not assisting or augmenting the user's control signal; rather, the user wants to make use of the robot's behavior as they understand it. Because of this, a robot's behavior should remain predictable under the partial control of the user. 

Legible robot motion refers to robot actions that are straightforward for a human to anticipate and comprehend. A common way to generate legible motion in goal-based robotic tasks is to model the user as having an internal cost function that is minimized when the robot's motion saliently moves towards a given goal \cite{dragan_generating_2013, faria_understanding_2021}. An alternative approach is to learn from humans through demonstrations or feedback \cite{busch_learning_2017, bied_integrating_2020}. Importantly, legibility and predictability are in the context of the robot \emph{completing the task} and are often in real-time as opposed to pre-hoc or post-hoc explanations \cite{das_explainable_2021, cruz_explainable_2021, sakai_explainable_2022, paleja_utility_2021}. Predictability is also an important part of our work. We operate under the assumption that a robot's behavior is predictable if the user has previously encountered similar behavior. This assumption is somewhat analogous to robot-centric concepts of out-of-distribution (OOD) states and behavior.

OOD detection is useful in many robotic and machine learning tasks \cite{yang_generalized_2022, sun_out--distribution_2022}. Identifying scenarios that are OOD relative to a robot's training data or past experiences can provide implications about the robot's environment or its performance, such as when an RL agent may behave sporadically or unexpectedly when in a state it has never been in before \cite{lan_can_2022}. It has also been used to identify when a robot may require feedback from a person to help complete a novel task \cite{dass_pato_2023}. In RL. OOD has recently been studied to infer when an agent is acting in a new MDP \cite{haider_out--distribution_2023}. We use similar techniques to detect when the robot is in a state that it would not otherwise act in. This can happen when a user is partially controlling a robot to perform a new task.

\section{Problem Setting}
\vspace{-1mm}
We describe a problem setting in which a user is accustomed to the autonomous execution of a task and wishes to partially control a robot during that execution to accomplish another task. The user creates a plan to partially control the robot based on how they expect the robot to behave to accomplish a novel task. Thus, it is critical that the robot behaves in a way that is predictable to the user, no matter the robot's current state. Given the robot behaves in a user-predictable way, as opposed to sporadically or in an unfamiliar way, a user can perform various novel tasks with little surprise and relative ease. 

In this setting, the user has seen the robot complete its task many times. We refer to this as a history of task ``rollouts.`` Based on this, we assume the following: when the robot is in a state the user has never seen before, they expect that the robot will act the same as it would do in the ``closest`` state to its current state. The ``closest`` is both problem and user-specific; however, the intuition is that the robot will behave similarly in similar circumstances and that in novel circumstances, a user will project onto what they have seen before. This assumption temporarily constrains the problem space. However, it is a reasonable assumption in many robotic tasks. Thus, the problem can be defined as: for any given state unseen to the user, the robot should find a state that the user has seen before and act as if it were in that state.   

We will now define the original task the robot can complete autonomously, and how this task is used to build up a user's expectation of the robot's behavior. Let task $orig$ be defined as an MDP with states $S\subset \mathbb{R}^n$, actions $A$, reward function $r: S \times A \rightarrow \mathbb{R}$, and transition function $T: S \times A \rightarrow S$. There is a robot that has learned an optimal policy for the task denoted $\pi^*$. Let $D$ be defined as a history of rollouts under $\pi_{orig}^*$ that the user has seen. Then, let the user's expectation of the robot's behavior given $D$ be $W_{D}:S \rightarrow S$. Here, $W$ is a function that maps from the robot's current state to the user's anticipation of what the next state will be.

To this setting, we introduce Partitioned Control (PC), where the user teleoperates one or more parts of the robot's actions. We separate the action space into two separate sets $A_U, A_R \subset A$; $A_U$ denotes the actions that the user can take and $A_R$ denotes the actions that the robot can take. We further assume that the user and robot action spaces are \emph{disjoint}; that is, $A_U \cap A_R = \{\mathbf{0}\}$. In other words, the user and the robot control different parts/different axes of the action space. For example, if the robot is acting in Cartesian space, the user may take control over x-axis actions, or take control over the rotation of a specific joint. We denote the user's expectation of how the robot will act with their partial control signal as $W_{D, U}:S \rightarrow S$, where $U$ is the user control. For brevity, hereafter we refer to this only as $W.$ 

To make the robot's behavior more predictable for the user, we want to adjust the behavior of the robot policy when it is outside the user's observation set $D$. The goal is to identify when the robot is in a novel state $s$ where there exists a state $s' \in D$ that leads to more predictable behavior. Formally, identify when $\exists s' \in D$ s.t.:
 \begin{equation}
     d(W(s), T(s, u \circ \pi^*(s))) \geq d(W(s), T(s, u \circ \pi^*(s')))
 \end{equation}
 where $d$ is a task-dependent distance metric between states, and $u\circ\pi^*(s)$ denotes the disjoint combination of the autonomous action of the robot and the user's teleoperation. When such a $s'$ is identified, the robot should act as if it were in $s'$. Specifically, we want to select a new proxy state $s'$ such that the user's \emph{predicted} state $W(s)$ is closer to the actual resultant state when simulating the policy in $s'$, $T(s, u \circ \pi^*(s'))$  than to the resultant state of running the policy directly $T(s, u \circ \pi^*(s))$. Lastly, in this setting, the true $W$ and the nature of the new task that the user wishes to accomplish are unknown. However, formalizing $W$ as such can be useful for modeling and/or simulating, creating a learning objective, or creating metrics to measure the success of algorithms applied to this problem.



 \begin{figure*}
\captionsetup[subfigure]{justification=centering}
    \centering
      \begin{subfigure}{0.3\textwidth}
        \includegraphics[width=\textwidth]{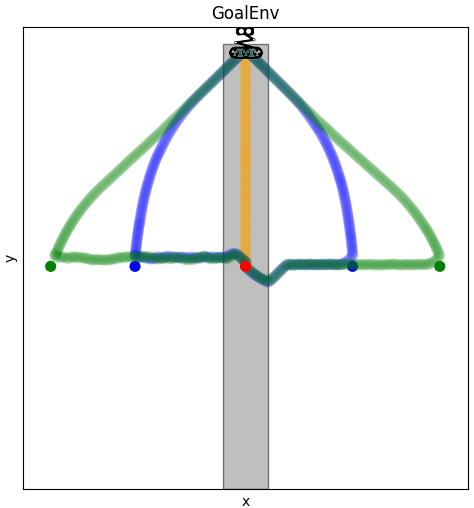}
          \caption{IODA \\ \phantom{a}}
          \label{fig:NiceImage1}
      \end{subfigure}
      \begin{subfigure}{0.3\textwidth}
        \includegraphics[width=\textwidth]{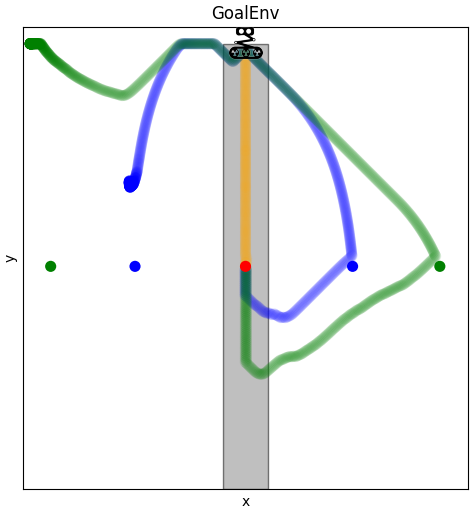}
          \caption{Out-of-distribution, \\ no imagined state}
          \label{fig:NiceImage2}
      \end{subfigure}
      \begin{subfigure}{0.3\textwidth}
        \includegraphics[width=\textwidth]{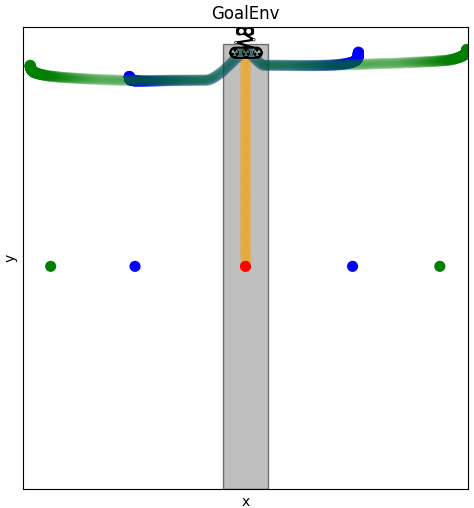}
          \caption{Must return to workspace, \\ no imagined state}
          \label{fig:NiceImage3}
      \end{subfigure}
      \vspace{-2mm}
      \caption{In a 2D goal navigation task, a simulated user is trying to leverage an optimal policy to reach subgoals by controlling the x-axis of the robot whilst the policy controls the y-axis. These subgoals are outside the robot's original workspace (highlighted in gray). Our algorithm \algo{} allows the user to seamlessly reach the subgoals.}
      \vspace*{-6mm}
\end{figure*}

\section{\algofull{} (\algo{})}
\vspace{-1mm}
In this section, we present \algofull{} (\algo{}), to facilitate a user to accomplish new tasks given a policy and a means of teleoperating an axis of robot behavior (the problem setting described in the previous section). Our key insight is that when the robot is acting in a region that greatly differs from what the user has seen before, the policy should act with imagined states that are as similar to the real state as possible while being ''in-distribution'' of what the user is familiar with and anticipates. Unless otherwise specified, we refer to ''in/out of distribution'' states with respect to $D$. 

\begin{algorithm}[t]
\caption{\algofull{} \\ (\algo{})}
\label{alg:algorithm}
\textbf{Initialize:} Rollout history D \\
\textbf{Initialize:} OOD-state detector \\
\textbf{Initialize:} State $s$
 
\begin{algorithmic}[1] 
\WHILE{not done}
\IF {s is OOD}
\STATE $s \rightarrow \arg\min_{s'\in D}d(s,s')$ 
\ENDIF
\STATE $a \rightarrow \pi^*(s)$
\STATE $u \rightarrow $ user's control signal
\STATE $s \rightarrow T(s, u\circ a)$
\ENDWHILE
\end{algorithmic}
\end{algorithm}

The complete \algo{} algorithm is presented in Algorithm 1. Here, we require that an OOD detector be trained on $D$. This is then used to detect ``novel`` states. While this technique is not new from a robot-centered perspective, it is also a human-modeling choice that draws an analogy between when a state is OOD and when a human may be projecting to a state they have seen in the past. Thus, it is also being used to determine when to search for a state that the robot policy should ``imagine`` it is in. 

\vspace{-2mm}
\section{Simulation Example}
\vspace{-1mm}

In this section, we demonstrate \algo{} in a 2d navigation task. In the original task (see Fig. 2), the robot learned to navigate to any specified goal point from within its workspace (highlighted in gray in Fig. 2). A user then seeks to leverage the behavior they have observed to control the x-axis actions to first guide the robot to an intermediate subgoal outside of the workspace (shown in Fig. 2 as blue or green dots) and then to the primary goal: this represents a novel task not represented by the robot's policy.

We train two RL agents to optimally solve two slightly different versions of the navigation task using the off-policy RL algorithm SAC \cite{haarnoja_soft_2019}. We use SAC as it has been shown to be relatively robust out of distribution \cite{lan_can_2022}. In one version of the task, Fig. 2, b, the RL algorithm was restricted to the gray workspace by being penalized for leaving it. In the other version, Fig. 2, c, the agent is penalized if it is out of the workspace, and further penalized by moving in the y-axis whilst it is out of the workspace. This encourages the agent to return to its workspace as quickly as possible before continuing the task. In both cases, the agent, when out of its workspace, may behave in a way unpredictable to a user. As a user has only seen optimal rollouts, they may not be familiar with what happens when the robot ``fails`` or is OOD and will likely expect that the robot would continue toward the primary goal along the y-axis.

In these environments, we collected 1000 rollouts of the optimal policy and trained Deep SVDD OOD detectors \cite{ruff_deep_2018} on the states of those rollouts. We choose $d$ to be the $L1$ distance between two states. Finally, we substitute human user control for an optimal x-position controller given the current x-position and subgoal location. As can be seen in Fig. 2. \algo{} is the only condition in which the simulated user can reach all subgoals and then easily proceed to the primary goal. In Fig. 2, b, the agent acted relatively sporadically when brought out-of-distribution, and could only reach both goals half the time.  In Fig. 2, c, since the agent was trained not to move in the y-axis when outside of its workspace, the agent's behavior inhibited the simulated user from reaching the subgoals. Furthermore, $D$ did not contain any indication that the robot would stop.

\section{Methodology: User Study}
\vspace{-1mm}


To study how users can leverage PC to accomplish new tasks as well as the efficacy of the IODA algorithm, we conducted an in-person user study, where people use PC with various underlying algorithms to accomplish a novel task. We hypothesize that users can leverage their expectations of robot behavior along with PC to accomplish this task. The IODA algorithm was designed to facilitate this. Thus we seek to validate users can use PC in this way and that when the robot's behavior more closely aligns with user expectations, the user can more readily complete the task.

\textbf{Plant Watering Task}
To study PC and IODA, we choose to replicate the scenario discussed in Section I. In this task, there is an RL robot policy that transports cups of liquid from one place to another, for handover, table setting, etc. The user then posits that they can use this task to water their flowers if they can rotate the robot's wrist as the robot carries the liquid to pour it over the flower bed. The fully autonomous component is the robot traveling along one side of the flower bed to the other, while users are prompted to pour out liquid to water the flowers by rotating the robot's wrist. This task is intuitive and entails PC over a single action space dimension and is thus suited for a study where participants are still relatively novice at teleoperation. 

The base policy for this task was trained using RL via SAC \cite{haarnoja_soft_2019}. The reward function used penalizes the robot per time step while the cup is not at the goal or if the robot spills liquid (by overly rotating the wrist or moving too fast). There is a large positive reward for reaching the desired goal position. Based on the setup for this task, rollouts of the optimal policy would not include the robot spilling or largely rotating its wrist past a threshold. However, this is precisely what users will need to do to perform the pouring task. While users have seen rollouts of the optimal policy they have not seen the robot train nor know what happens when the robot enters the OOD state of pouring out liquid.

\textbf{Experimental setup}
The setup consisted of a Kinova Gen3 7-dof arm located between two tables (Fig. 1). A table was used for the participant to practice controlling the robot through teleoperation; the other was used to demonstrate the robot cup-carrying policy and for the watering task. The pouring material used were small beads meant to replicate pouring a fluid. The flower bed was equipped with 5 different containers, each of which had scales below them to measure the amount of beads poured into each container. Each container represents an individual flower. 

\vspace{-2mm}
\subsection{Conditions}
\vspace{-1mm}
For all conditions, the policy for the robot's autonomous behavior is constant. The robot will attempt to transport a cup full of beads from one end of the flower bed to the other. While the robot is doing this, the user will have roll axis control over the robot's wrist. Each condition is an approach one may take for PC.

\textbf{Unaltered Base Policy}
In this condition, there is no alteration to the base liquid carrying policy during the user's PC. As the setting and PC are novel, this intuitive baseline is important to serve for both studying a user's experience during PC and how an underlying policy may perform in these scenarios. However, because this policy is unaltered, it may suffer from the problems associated with out-of-distribution states examined earlier. That is, the robot may act or move sporadically along its path if the user's control causes the robot to start spilling the beads (which they are intentionally trying to do). We expect that this will result in both lower task success and that the robot's behavior will not align with the user's expectations based on what they have seen prior. We will refer to this as the \textit{RL} condition.

\textbf{Base Policy with Enhanced Failure Recovery ("STOP")}
In this condition, the base policy also has an explicit failure recovery component. Although the unaltered policy may still try to recover from spilling liquid, there is no explicit instruction for what the robot should do while spilling. For example, the robot should stop moving along its path to minimize the spread of the spill. In this condition, however, there is an additional safety constraint that while the robot is spilling, it will stop moving along its path until it is no longer spelling liquid. This behavior is likely desirable for a "carry liquid policy," but it also may or may not be expected by a user who has only seen successful policy examples. We expect that, especially for users who do not expect this stopping behavior, many of the beads will accidentally be poured into one or two flowers as opposed to an even spread. This is because, if the stopping behavior is unexpected, a user may need time to react to the robot stopping moving along its path once the beads start pouring out. We will refer to this as the \textit{STOP} condition.

\textbf{IODA}
In this condition, we apply the IODA algorithm while the user is engaged in PC. An OOD detector was trained on optimal policy rollouts before the study. We used the L1 distance as the distance function used in the algorithm. Because the IODA algorithm will result in the robot roughly following its original path regardless of the presence of PC, we expect that as users rotate the robot's wrist, beads will be evenly poured into the flower basin. Furthermore, this is the behavior we hypothesize that users will expect. 

\subsection{Experimental procedure}
After participants read and signed an informed consent form, they practiced teleoperating the robot for up to three minutes. For practice, users were given XYZ control as well as roll/wrist control as the robot grasped an empty cup and were encouraged to get comfortable controlling the robot. The speed of roll rotation matched the speed during each condition. The purpose of this practice task was to ensure that all users had a similar minimum level of familiarity with controlling the robot before moving on to the pouring task. 

After the practice session, we explained that the robot had an autonomous policy to carry cups of liquid. Users then watched the robot carry a cup of beads to three different locations. We will refer to this as the familiarization phase. After familiarizing themselves with the robot's behavior, they were then instructed that they would take control of the robot's wrist as it carried a cup of beads from one end of a flower bed to another. Their task was to try to water the five flowers as evenly as possible while using the most beads possible. Participants would then complete the pouring task in one of the 3 conditions (the choice of which was fully counterbalanced). The task ended after one minute or until the robot reached its goal pose, irrespective of how many beads they had already successfully poured. Users completed a post-condition survey after their experience. We repeated this in each of the two remaining conditions. Finally, users were thanked and given compensation.

\textbf{Outcome Measures}
The post-condition survey included questions from the UTAUT \cite{venkatesh_user_2003} survey. We adjusted the scale of all questions to a 5-item Likert-scale. We also asked two other Likert-scale questions: \emph{How much did the robot’s behavior align with your expectations?} and \emph{I was surprised by the robot's behavior}, and two free-response questions: \emph{Did the robot behave as you expected? If not, please explain how. } and \emph{How much do you feel the robot's ability to complete the task depended on your input?}. 

For a quantitative performance metric, we define ''pour error." We measure the total deficit of the bins relative to an optimal pour of $w = 68 \text{g}$ each. We measure the deficits and not the overfills since measuring overfill would count this error twice. This pour error $\phi$ is equivalent to measuring the total amount of beads lost in the process, combining bin overflow, beads that did not land in bins, and beads remaining in the cup. Thus,
\vspace*{-3mm}
\begin{equation}
    \phi = \sum_{i=1}^5 \max(w - b_i, 0),
\end{equation}
where $b_i$ represents the measured weight of beads in bin $i$.


\textbf{Hypotheses}
Based on what we know about how the robot will act under PC in each of the three conditions, we propose the following hypotheses.  \textbf{H1}: IODA will most meet user expectations, followed by STOP and then RL; \textbf{H2}: IODA will lead to overall the best task performance, followed bt STOP and then RL; \textbf{H3}: In PC, there will be a strong positive correlation between meeting a user's expectation and task performance.

\begin{figure}[t]
\centering
\includegraphics[width=.475\textwidth]{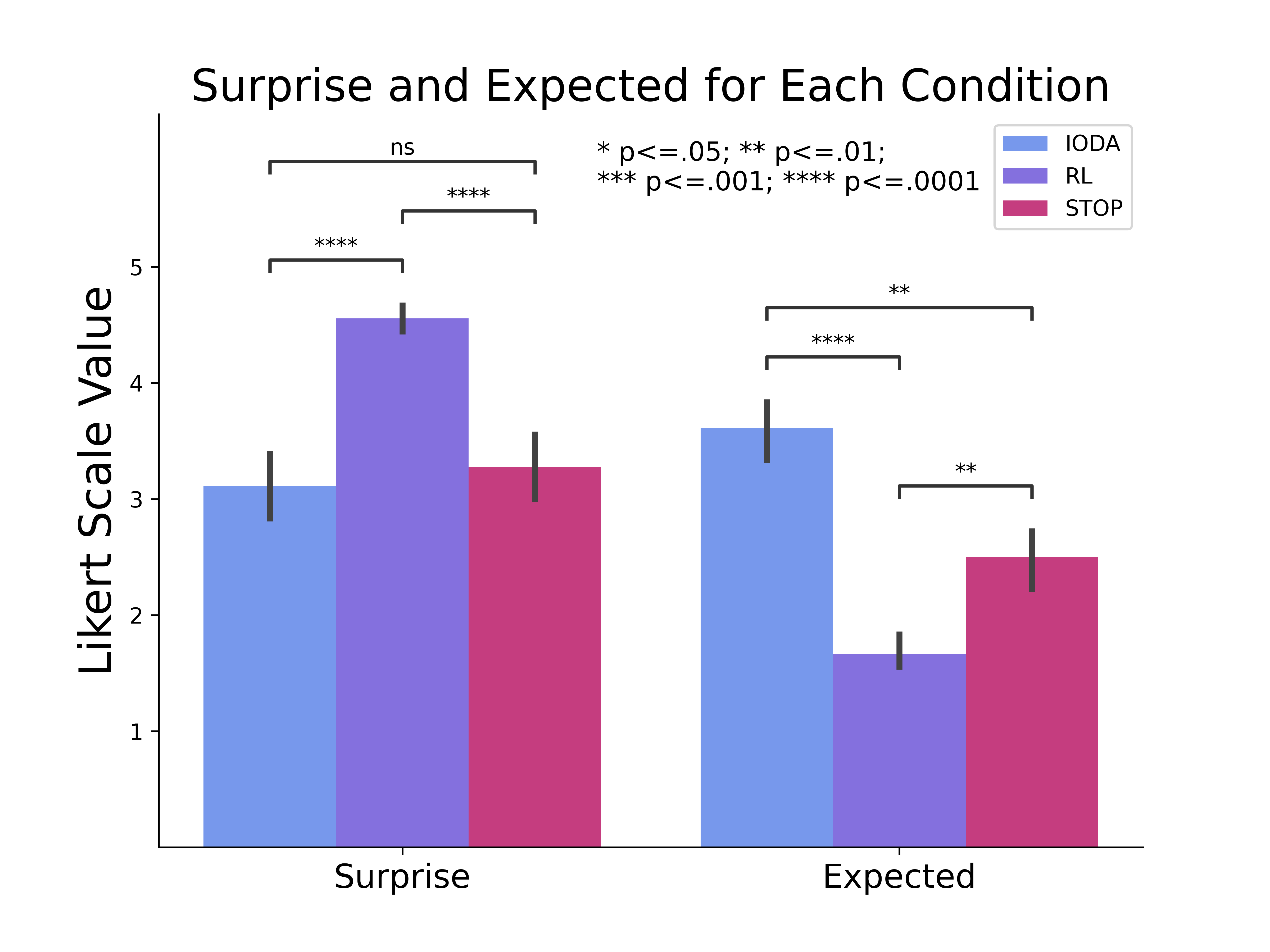}%
\vspace*{-5mm}
\captionsetup{
indention=1.5em
}
\captionof{figure}{User reported expectation alignment and degree of surprise for each condition.
\vspace*{-8mm}}%

\end{figure}

\vspace{-1mm}
\subsection{Results}
\vspace{-1mm}
 \textit{Participants} We recruited 18 participants from the university and the surrounding area with a variety of different backgrounds. All participants were 18 years or older. Of these participants, 10 were female, 6 were male, 1 was nonbinary, and 1 was genderqueer. 13 participants were in the 18-24 age range, 4 in the 25-35 age range, and 1 in the 35-44 age range. Of these participants, 2 were self-reported robot experts (i.e., attend robotics conferences regularly), while all other participants reported having interacted with a robot in the past (i.e., a Roomba). The study lasted approximately 30 minutes and participants were compensated $\$10$.  The study procedure was approved by the Tufts University IRB. 

 \textit{Analysis} To analyze the data we use both Bayesian statistics and p-values. All tests were done using independent samples t-tests and the Bayesian tests were done with a Cauchy prior distribution with $r=1/\sqrt{2}$.


\textit{User Expectations} Before watering the flowers in any of the three conditions, as mentioned, users had both practice time and were able to watch the robot carry cups of beads to familiarize themselves with its movement. We expect these two initial phases, as well as any ordering effect, influenced how a user reported both how much the robot's behavior met their expectations and how surprised they were by the interaction. The results of the postcondition Likert-scale questions can be found in Fig. 3. As we can see, IODA led to robot behavior that both best aligned with people's expectations and induced the least amount of surprise. Specifically, IODA met user expectations to a greater extent over RL (\textit{p} $\approx$ 0.0, \textbf{BF} $>$ 10000), and to a slightly greater extent over STOP (\textit{p} $\approx$ 0.0064, \textbf{BF} = 7.05). Comparing STOP to RL, we find STOP more closely meets user expectations (\textit{p} $\approx$ 0.0099, \textbf{BF} = 5.035). A large part of why the RL condition least met user expectations is because the sporadic behavior caused by the RL policy being out-of-distribution when the robot's wrist was rotated was that it would begin to move away from the participant as opposed to towards and away from the center of the flower bed (Fig. 5). These results support \textbf{H1}.

\begin{figure}[ht]
    \begin{subfigure}{.475\textwidth}
    \centering
    \includegraphics[scale=0.38]{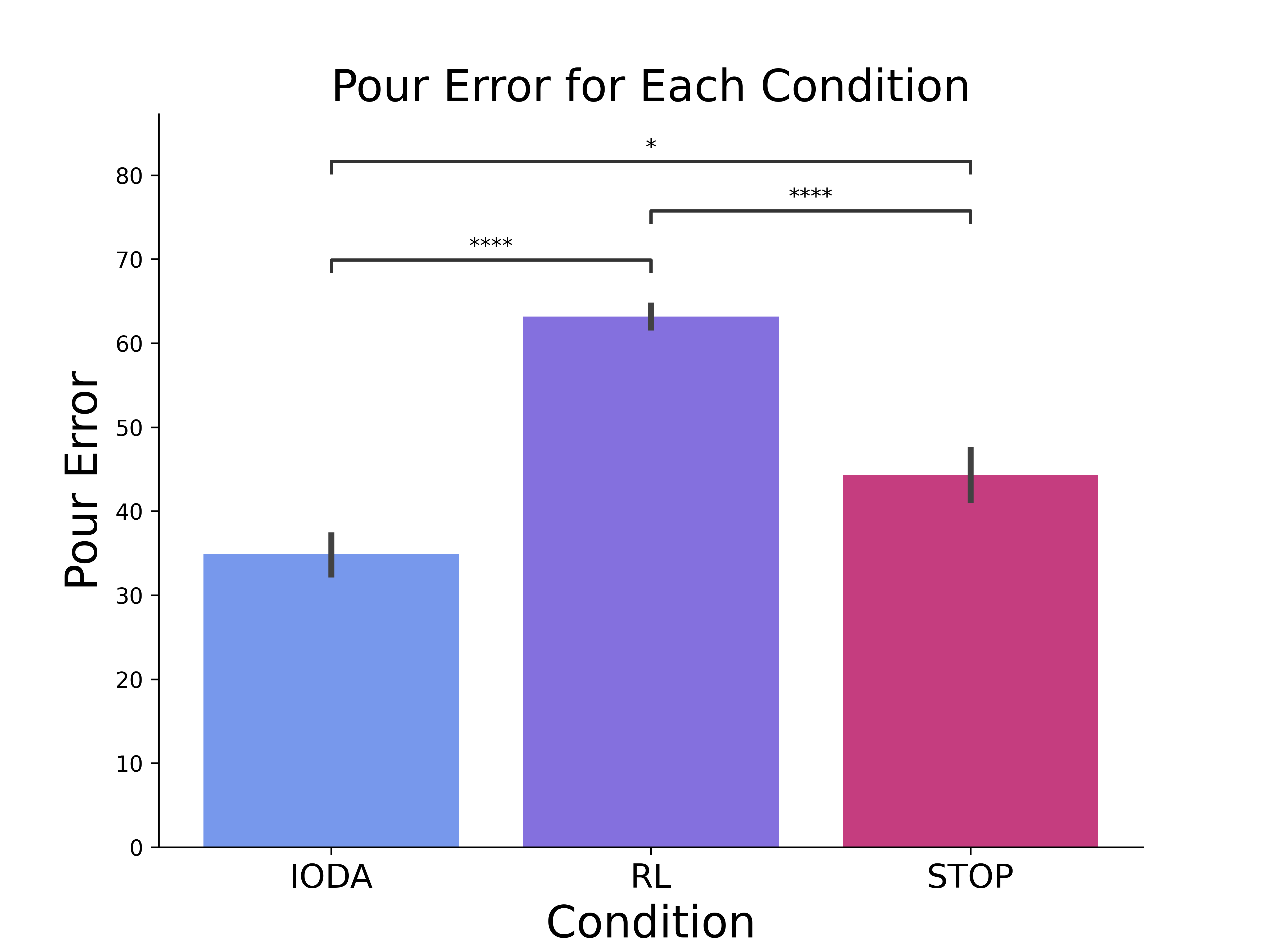}
    \label{subfig:a}
    \vspace*{2mm}
    \end{subfigure}
    \begin{subfigure}{.475\textwidth}
    \centering
    \vspace{-1mm}
    \includegraphics[scale=.38]
    {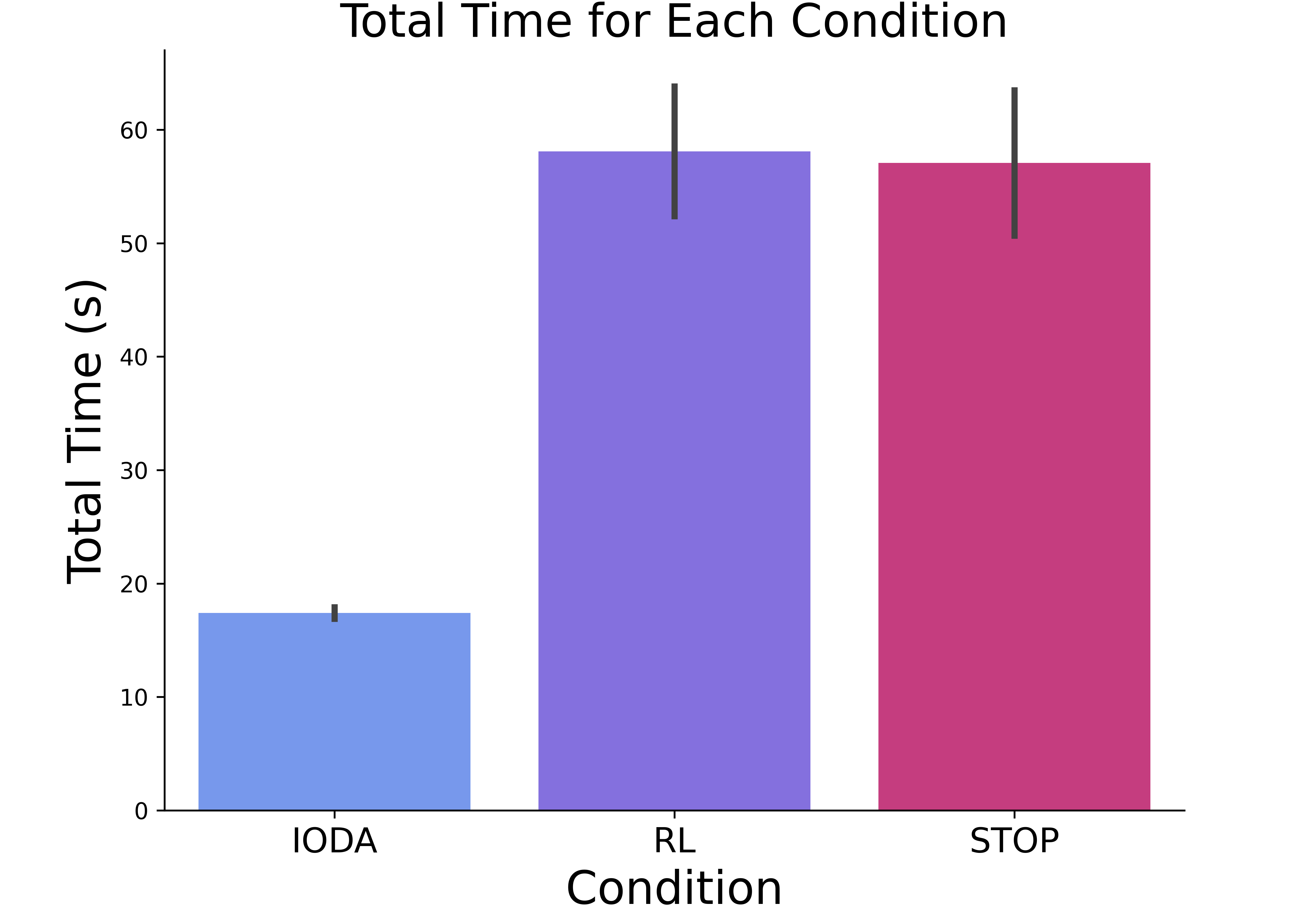}
    \label{subfig:b}
    \end{subfigure}
    \vspace{-2mm}
    \caption{\textit{Top}: IODA performed the best in the watering task with the least error. \textit{Bottom}: Mean and standard-deviation for time-on-task for each condition}\label{fig:post}
     \vspace{-4mm}
\end{figure}

\textit{Task Performance} The primary task metric we analyze is \textit{pour error}. The results are shown in Fig. 4. We find IODA led to much better task performance than RL (\textit{p} $\approx$ 0.0, \textbf{BF} $>$ 10000) and slightly better performance than STOP (\textit{p} $\approx$ 0.020, \textbf{BF} = 2.974). Notably, in the STOP condition, many users reported "figuring it out" after some trial and error. This is partially captured in the time-on-task chart in Fig. 4, although even an expert in the STOP condition would still take longer than in the IODA condition due to the nature of the stopping behavior. That being said, we do find IODA led to slightly better performance with significantly less time-on-task. Similarly, 5 of 18 participants did figure out that in the RL condition, they could wait for the cup to be almost at the end of the flower bed and then begin rotating the robot's wrist so that it would move back across the flower bed while pouring out the beads. However, this took most of the participants almost the entire 60-second trial time to realize. These results support \textbf{H2}.

\begin{figure*}[t]
  \includegraphics[scale=.18, width=\textwidth]{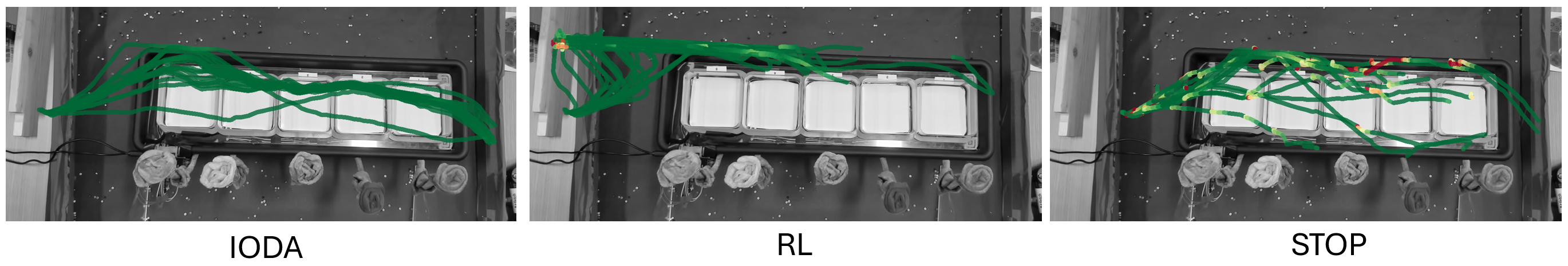}
  \vspace{-6mm}
  \caption{Trajectories of the cup for all 18 participants. The redder the line indicates how long the cup was stopped at that point. The reddest point indicates that the cup is stopped for at least 7.5 seconds}\label{fig:study}
  \vspace*{-6mm}
\end{figure*}

\begin{figure}[]
    \begin{subfigure}{.475\textwidth}
    \centering
    \includegraphics[scale=0.4]{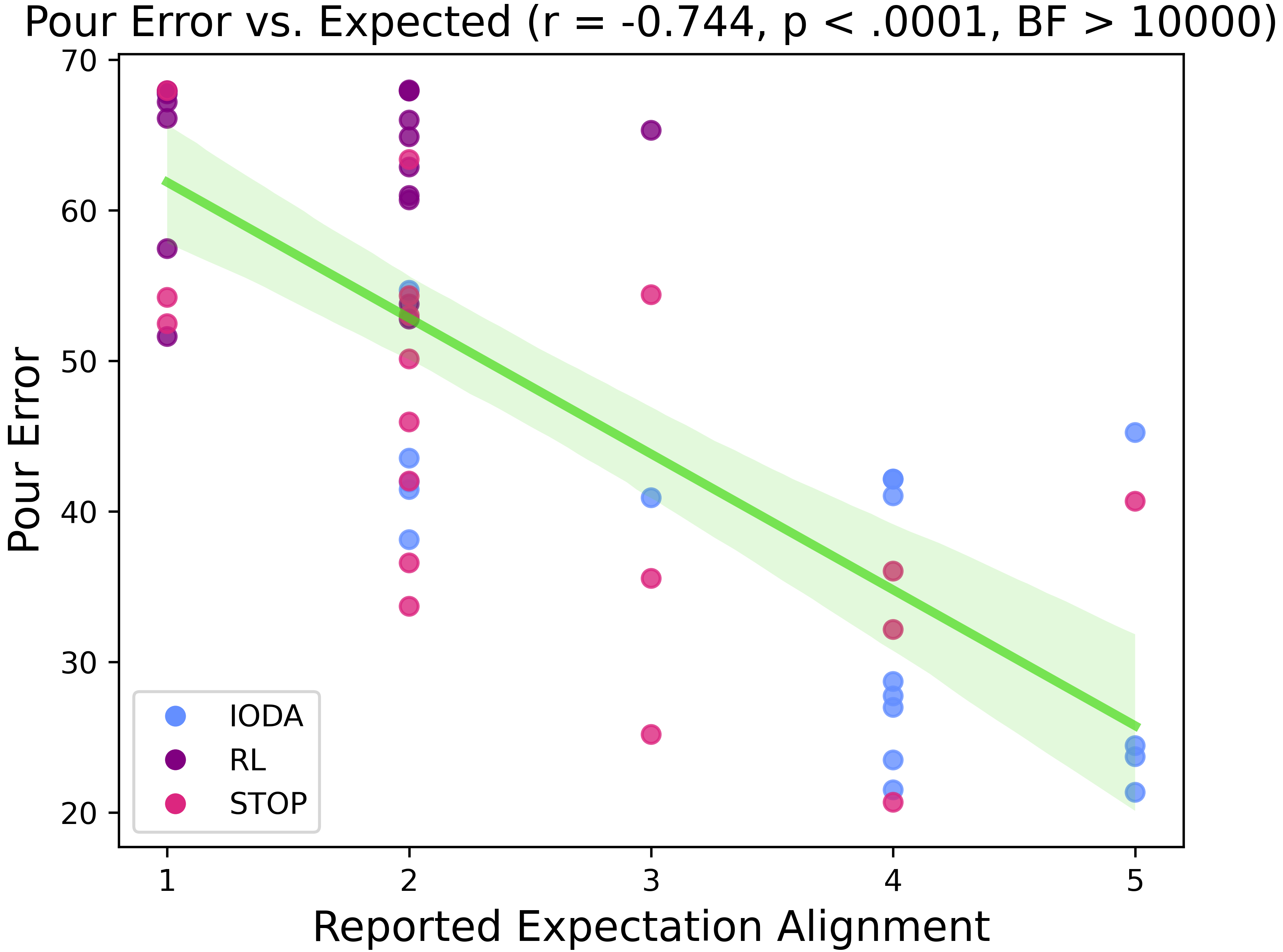}
    \label{subfig:a}
    \vspace{1mm}
    \end{subfigure}
    \begin{subfigure}{.475\textwidth}
    \centering
    \includegraphics[scale=.4]{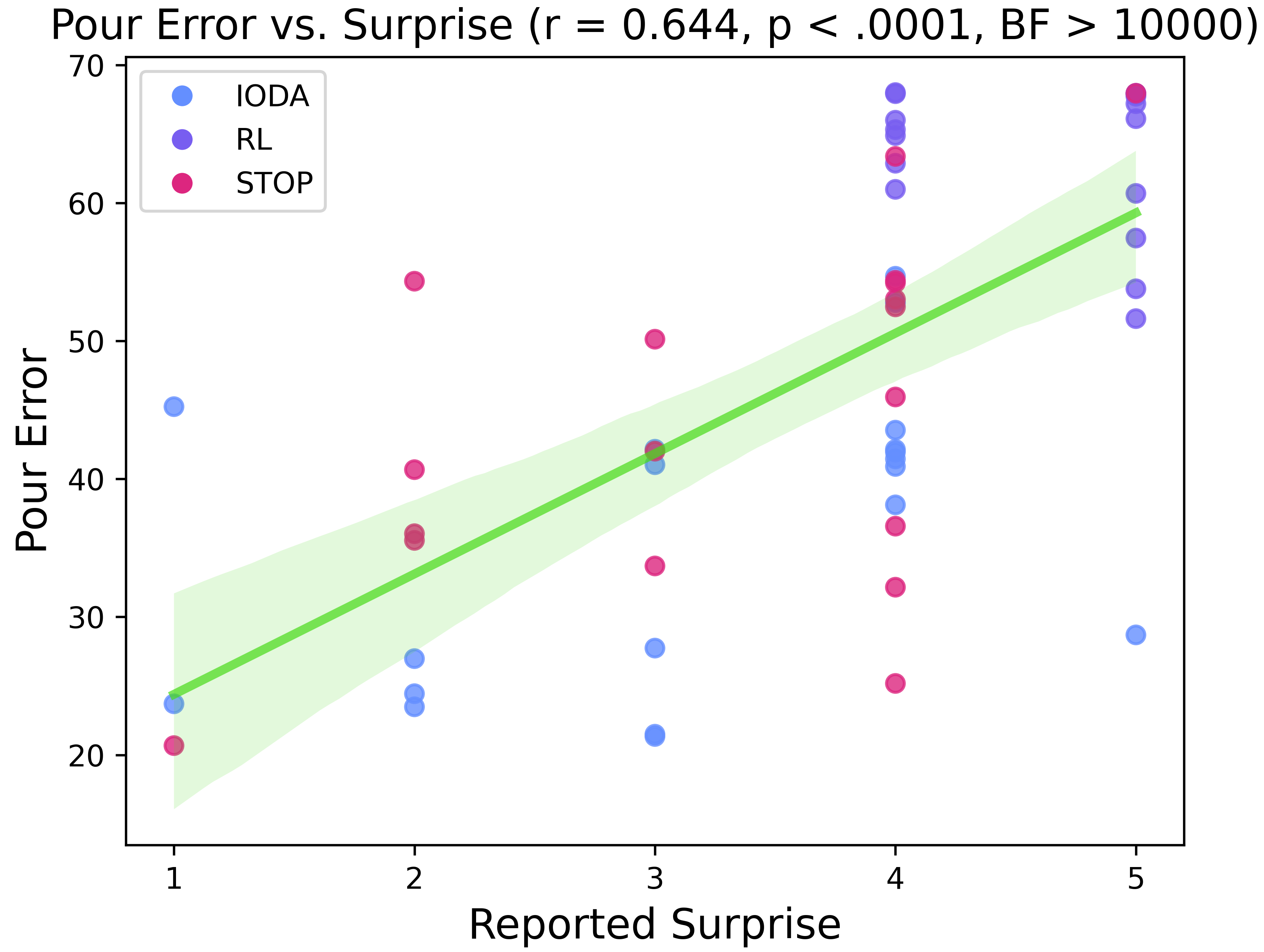}
    \label{subfig:b}
    \end{subfigure}
    \vspace{-1mm}
    \caption{\textit{Top}: Meeting user's expectations is strongly correlated with task performance in PC. \textit{Bottom}: The same is true of reducing surprise and performance.}\label{fig:post}
     \vspace{-8mm}
\end{figure}

\textit{The Importance of Meeting User Expectations in Partitioned Control} We hypothesize that during PC, robot behavior that meets a user's expectations will correlate to higher task success. Although various collaborative shared control paradigms are designed to work despite a user's expectations or work under the assumption that a user and robot share a world model, as in classic SA \cite{javdani_shared_2015}, in PC, aligning user expectations and robot behavior is critical for task and user performance. We analyze the relationship between the user's reported expectation alignment and the user's reported surprise. The results are displayed in Fig. 6. We find that there is a strong correlation between high task performance (low pour error) and meeting user's expectations, as well as a strong correlation between low levels of surprise and high task performance. These results support \textbf{H3}. 

\vspace{-2mm}
\section{Discussion}
\vspace{-1mm}
A na\"ive learned robot policy may not be suitable for flexible interactions with real users, especially when they have the propensity to use the policy in unexpected ways. Such a propensity is not exclusive to human-robot collaboration: people will use a shovel as a crowbar, a crowbar as a hammer, a hammer as a hook, and a hook as a shovel.  We investigated Partitioned Control (PC), in which a user controls some dimensions of the behavior of an RL-trained robot and can use that control to drive it into states that are not reflected in training. We present an approach, Imagined Out of Distribution Actions (IODA) that enables such a partially-controlled system to behave in alignment with user expectations.  We demonstrated that a standard RL-trained agent will behave erratically under PC, while IODA results in more expected robot behavior.  Furthermore, we show that in a realistic PC setting, when a robot's behavior is more aligned with a user's expectations, the user can more effectively perform the novel task they are trying to achieve. 

There are, however, aspects of this which warrant further investigation. One is to study how users build up their expectations before and during PC. Here, we assumed that user expectations are based on teleoperation experiences and viewing prior rollouts of a given policy. However, there may be other important factors. A second aspect is the use of distance functions over the state to quantify user expectation alignment. We assumed that a distance function can be used as a proxy for what a user considers ``similar'' states, and we used an OOD detector to approximate when the robot is in a state that a user is unfamiliar with. While this approach was effective in our user study, there is more to learn about the properties and assumptions of IODA and the use of imagination to better meet user expectations. IODA may also run into latency issues if the calculations of the distance between states are not relatively fast. 

Our study addressed a ''one-shot" interaction where users performed the flower watering task with each condition once. This is an important setting because in many real-life scenarios, it is ideal for a task to work on the first go. Enabling ''one-shot'' interactions improves user satisfaction and generates successful demonstrations that could be used to learn the new task. That said, most users, across all three conditions and regardless of whether the robot met their expectations, wanted to interact with the robot again. This is not only because they enjoyed the task, but also because they thought they could better perform the task knowing what to expect. Thus, user expectations change as a result of PC interactions, and future work is needed to address how PC and IODA change these expectations over time.

\vspace{-2mm}
\section{Conclusion}
\vspace{-1mm}
In this paper, we studied the scenario in which a user
wishes to leverage their knowledge of a robot policy to perform a novel task. Users may want to do this through partitioned control, controlling one or more axes of a robot's actions while the others remain autonomous. However, an RL policy may not facilitate this interaction without ending up in an OOD state and potentially acting unexpectedly. IODA takes into account a user's prior interactions with the robot and leverages imagined states to better meet user expectations, leading to more successful task outcomes. 

\section*{ACKNOWLEDGMENT}
\vspace{-2mm}
The work described here was supported in part by the US
National Science Foundation (IIS-2132887)

\bibliographystyle{IEEEtran}
\bibliography{references}
\end{document}